\pdfoutput=1
\documentclass[conference]{IEEEtran}
\IEEEoverridecommandlockouts
\usepackage[T1]{fontenc}
\usepackage{tgtermes}
\usepackage[numbers,sort&compress]{natbib}
\usepackage{graphicx}
\usepackage{booktabs}
\usepackage{array}
\usepackage{xcolor}
\usepackage{amsmath,amssymb,mathtools,amsthm}
\usepackage{xurl}
\usepackage[hidelinks]{hyperref}

\theoremstyle{plain}
\newtheorem{theorem}{Theorem}
\newtheorem{proposition}[theorem]{Proposition}
\newtheorem{lemma}[theorem]{Lemma}
\theoremstyle{definition}

\newcommand{\cfConfCellCapxferLoneCfown}{51.8\,\%}
\newcommand{\cfConfCellCapxferLoneEdecown}{0.99\,\%}
\newcommand{\cfConfCellCapxferMergeEdecown}{16.47\,\%}
\newcommand{\cfConfCellCapxferNeps}{210}
\newcommand{\cfConfCellCapxferStratumCfown}{84.7\,\%}
\newcommand{\cfConfCellCapxferStratumEdecown}{3.22\,\%}
\newcommand{\cfConfCellOursLoneCfown}{72.8\,\%}
\newcommand{\cfConfCellOursLoneEdecown}{0.12\,\%}
\newcommand{\cfConfCellOursMergeEdecown}{0.35\,\%}
\newcommand{\cfConfCellOursMultiEdecown}{1.03\,\%}
\newcommand{\cfConfCellOursStratumCfown}{95.3\,\%}
\newcommand{\cfConfCellOursStratumEdecown}{0.21\,\%}
\newcommand{\cfConfExplainDeficit}{77.1\,\%}
\newcommand{\cfConfExplainFalsereason}{13.0\,\%}
\newcommand{\cfConfGateParUndom}{0.49\,\%}
\newcommand{\cfConfGateRawUndom}{97.1\,\%}
\newcommand{\cfConfHetCfdirectHone}{-3.5}
\newcommand{\cfConfHetCfdropHtwo}{19.8}
\newcommand{\cfConfHetEepHone}{0.44\,\%}
\newcommand{\cfConfHetEepHtwo}{3.2\,\%}
\newcommand{\cfConfMainEdecMatched}{3.2\,\%}
\newcommand{\cfConfMainEep}{0.95\,\%}
\newcommand{\cfConfMainKindLoneMatched}{2.9\,\%}
\newcommand{\cfConfMainKindMergeMatched}{4.2\,\%}
\newcommand{\cfConfMainKindMultiMatched}{6.7\,\%}
\newcommand{\cfConfMainNeps}{1,050}
\newcommand{\cfConfMainPOne}{38.5}
\newcommand{\cfConfMainPTwo}{35.3}
\newcommand{\cfConfMergeCf}{90.8\,\%}
\newcommand{\cfConfMergeFlipTOne}{0.07\,\%}
\newcommand{\cfConfMergeFlipTTwo}{0.13\,\%}
\newcommand{\cfConfMergeMiscovTOne}{0.41\,\%}
\newcommand{\cfConfMergeMiscovTTwo}{2.87\,\%}
\newcommand{\cfConfMergeNpairs}{975}
\newcommand{\cfConfMergePFour}{55.2}
\newcommand{\cfConfPosthocCalsd}{3.0}
\newcommand{\cfConfPosthocMiscov}{12.5\,\%}
\newcommand{\cfConfPosthocNcommon}{40}
\newcommand{\cfConfPosthocPOne}{46.6}
\newcommand{\cfConfPosthocPTwo}{34.4}
\newcommand{\cfConfScaleCovNine}{81.2\,\%}
\newcommand{\cfConfScaleCovSeven}{85.8\,\%}
\newcommand{\cfConfScaleEepNine}{1.7\,\%}
\newcommand{\cfConfScaleEepSeven}{1.7\,\%}

\newcommand{\cfSaDiagUndom}{99.9\,\%}

\newcommand{\cfTFourDregretEvoiClamp}{-6.0}
\newcommand{\cfTFourDregretEvoiDominance}{-1.9}
\newcommand{\cfTFourDregrethiEvoiClamp}{-4.9}
\newcommand{\cfTFourDregretloEvoiClamp}{-6.9}
\newcommand{\cfTFourEpunsafeDominanceClamp}{0.3\,\%}
\newcommand{\cfTFourEpunsafeEvoi}{0.3\,\%}
\newcommand{\cfTFourEpunsafeFixedVTwo}{0.2\,\%}
\newcommand{\cfTFourEpunsafeFixedVTwoClamp}{0.3\,\%}
\newcommand{\cfTFourRegretDominanceClamp}{18.8}
\newcommand{\cfTFourRegretEvoi}{16.9}
\newcommand{\cfTFourRegretFixedVTwo}{31.8}
\newcommand{\cfTFourRegretFixedVTwoClamp}{22.9}
\newcommand{\cfTFourUsefulDominanceClamp}{93.3\,\%}
\newcommand{\cfTFourUsefulEvoi}{93.1\,\%}
\newcommand{\cfTFourUsefulFixedVTwo}{82.7\,\%}
\newcommand{\cfTFourUsefulFixedVTwoClamp}{93.4\,\%}

\newcommand{\cfTeaserBodies}{42}
\newcommand{\cfTeaserPool}{236}

\newcommand{\cfProbeExtraDominance}{4.5}
\newcommand{\cfProbeExtraEvoi}{1.7}
\newcommand{\cfProbeShareEvoi}{19\,\%}
\newcommand{\cfScreenHosEep}{1.1\,\%}
\newcommand{\cfScreenHosNeps}{267}
\newcommand{\cfScreenHosUseful}{88.5\,\%}
\newcommand{\cfScreenKneeEep}{0.3\,\%}
\newcommand{\cfScreenKneeNeps}{585}
\newcommand{\cfScreenKneeUseful}{93.1\,\%}
\newcommand{\cfScreenNominalNeps}{585}
\newcommand{\cfScreenNominalUseful}{64.9\,\%}

\newcommand{\cfFigTwoShareClampNSixteen}{17\,\%}
\newcommand{\cfFigTwoShareVtwoNFour}{53\,\%}
\newcommand{\cfFigTwoShareVtwoNSixteen}{59\,\%}
\newcommand{\cfFigTwoUndomYawMultiClamp}{98.7\,\%}

\newcommand{\cfFigFourMergeMissesTOne}{8}
\newcommand{\cfFigFourMergeMissesTTwo}{72}

\newcommand{\cfFigFiveParSThreeEep}{1.2\,\%}
\newcommand{\cfFigFiveRawSThreeEep}{34.5\,\%}
\newcommand{\cfFigFiveRawSZeroEep}{0.61\,\%}

\newcommand{\cfFigconceptMaskG}{0.5}
\newcommand{\cfFigconceptMaskRhoAlone}{87.1\,\%}
\newcommand{\cfFigconceptMaskRhoPair}{98.5\,\%}
\newcommand{\cfFigconceptMaskRhoParent}{99.6\,\%}
\newcommand{\cfFigconceptMaskUneedParent}{1.72}
\newcommand{\cfFigconceptSpeedDiag}{0.86}
\newcommand{\cfFigconceptSpeedForward}{0.61}
\newcommand{\cfFigconceptSpeedModeltop}{0.92}
\newcommand{\cfFigconceptSpeedProbetop}{0.67}
\newcommand{\cfFigconceptSpeedTruetop}{0.77}

\begin{document}
\title{Which Configuration Can Do the Job?\\Certified Capability Prediction for Modular Robots That Split and Merge}
\author{\IEEEauthorblockN{Liam Kenji Bray}
\IEEEauthorblockA{Independent Researcher\\
liambray\_@outlook.com}
\and
\IEEEauthorblockN{Nicholas Cheng}
\IEEEauthorblockA{Independent Researcher\\
nicholasicheng6@gmail.com}}
\maketitle

\begin{abstract}
A modular robot team that splits and merges must know, before it reconfigures, which of its possible configurations can do a job. This is hard for two reasons. Docking hides weak modules, because healthy neighbours carry them, and a
merge with another team creates a configuration that has never been driven. We introduce \emph{certified capability
prediction}: from one 12-s test drive of the current assembly, give a lower bound on how well every possible child or merge can move in each direction. A statistical guarantee limits the chance of certifying a configuration for a job it cannot do (an \emph{unsafe commitment}). We show that the test drive observes only some of these answers; the rest are extrapolated through a model of the motors, and are only as right as that model. The certificate combines a learned estimator of each module's state, a physics simulation of each
candidate and one calibrated safety margin. In a pre-registered simulation study with mecanum-wheeled modules, only
\cfConfMainEep{} of test episodes contained an unsafe commitment, against a 10\,\% limit. At each method's own bound,
our certificate certified \cfConfCellOursLoneCfown{} of capable faulty modules left alone after a split, against
\cfConfCellCapxferLoneCfown{} for the best prior method, with fewer unsafe commitments. A compositional version
certified \cfConfMergeCf{} of capable merges of modules never docked together.
\end{abstract}

\begin{figure*}[t]
  \centering
  \includegraphics[width=\textwidth]{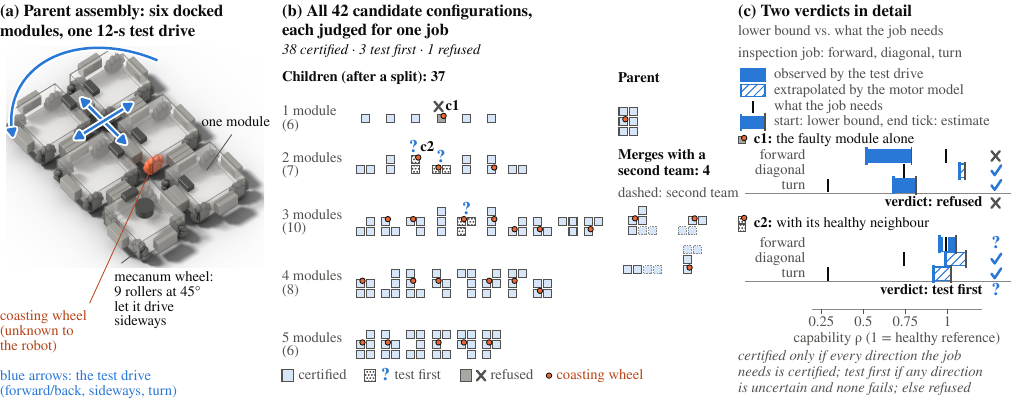}
  \caption{The question this paper answers. (a) Six docked modules, each a square platform with four mecanum wheels
(rollers at $45^\circ$ let a configuration drive in any direction). One wheel coasts, unknown to the robot. The assembly runs
one 12-s test drive. (b) From that test drive our certificate judges every configuration the team could become, here
\cfTeaserBodies: the assembly itself, every child it could split into, and every merge with another team. For one job,
each configuration is certified (its lower bound clears what the job needs), marked \emph{test first} (its estimate clears what the job needs but its lower bound does not, so a further 8.8-s test could settle it; Sec.~\ref{sec:problem}), or refused (even the estimate falls short). This illustrative job needs three directions, each with its own requirement: a configuration is certified only if every needed direction is certified, refused if any is refused, and marked test first otherwise. (c) Inside two verdicts: the lower bound and estimate of capability $\rho$ (defined in Sec.~\ref{sec:problem}) in each direction the job needs,
against the requirement. Hatched bars need wheel operating points (speed, torque) the test drive never covered, so their answer comes from the
motor model. Rendered from the module's real dimensions; one illustrative test episode of the pre-registered run
(selection rule in Appendix~\ref{app:prereg}).}\label{fig:teaser}
\end{figure*}

\section{Introduction}\label{sec:intro}
A modular robot team can reshape itself for each job, but only if it knows in advance which shapes will work. Such a robot is built from many copies of one unit, a \emph{module}, that can dock to its neighbours and undock again. Docked modules form one larger robot that moves and carries loads together; undocked, they become several
smaller robots that can cover more ground. Each connected group of docked modules that moves as one robot is a \emph{configuration}. The current one is the \emph{parent}; splitting it yields \emph{children}, and docking with another team a \emph{merge}. The promise of modularity is that a team can choose its configuration for each job.

Our modules are small square wheeled platforms (Fig.~\ref{fig:teaser}a; specifications in Table~\ref{tab:module}). Each has four \emph{mecanum} wheels, whose
rims carry small rollers mounted at $45^\circ$, so that a configuration made of them can drive forward, sideways, diagonally,
or turn on the spot by choosing each wheel's speed. 

Choosing a configuration for a job means answering one question before reconfiguring: \emph{which of the configurations we could
become can do this job?} A child robot sent on a job it cannot finish stalls halfway; a merged robot trusted with a
load it cannot move blocks the task for everyone. The simplest answer is to assume that every module is as
good as new. Real modules are not. Even modules built to one design differ: motors of the same model vary in torque and top
speed by several percent, and
battery charge shifts both; gearboxes wear and collect debris; a motor driver or connection can degrade so that a
wheel receives only part of its voltage; and a motor can fail open, leaving the wheel back-driven through its gearbox, or short, braking it.

Two facts make the question hard. First, an assembly hides these differences. A speed controller on every wheel keeps it at its commanded speed by setting the motor driver's PWM \emph{duty} $u\in[-1,1]$; a healthy motor then sees an average voltage $u$ times the battery voltage. When a module with a weak wheel is
docked to healthy ones, the weak wheel's controller turns its duty up, to full if need be, and the healthy wheels pull
it along, so the assembly still drives almost normally. Once the module is split off, alone or in
a smaller child, fewer healthy wheels share the work, and the shortfall appears (Fig.~\ref{fig:concept}a). Second, a merge with another team creates a
configuration that has never been driven, even when every module in it is healthy. Forming each candidate configuration just to test it would
cost a reconfiguration per candidate, and a merge would need the other team to travel first.

We therefore ask the robot to answer before the new configuration exists. We call this \emph{certified capability
prediction}: from one short test drive of the current assembly, give for every configuration it could form a lower bound on
how well that configuration can move in each direction, and commit a configuration to a job only where the bound clears what the job
needs. The bounds come with a guarantee: the chance that any commitment in an \emph{episode} (one test drive and every decision it informs) is unsafe is at most a chosen
level $\alpha$. The prediction is \emph{counterfactual}: it concerns configurations the robot is
not in.

Our central insight is that a test drive observes only some of the answers; the rest are extrapolated through a motor model, and a robot that does not tell the two apart trusts both equally. At a fixed terminal voltage, a brushed DC gearmotor's torque falls linearly with speed, from its stall torque at rest to zero at its no-load speed. A lower voltage shifts this line down in parallel. We call this torque--speed relation the \emph{motor model}. The test drive shows each wheel at a few combinations of speed and
torque. A new configuration sometimes needs its wheels at combinations the test never reached: driving diagonally, or a small child turning on
the spot, spins some wheels faster than the parent ever did. For those questions the answer does not come from data but
from extrapolating the motor model (Fig.~\ref{fig:concept}b). Such answers sound as confident as measured ones, yet they are only as right as the motor model.

This paper makes four contributions.
\begin{itemize}
\item (C1) \emph{A problem statement:} certified capability prediction for every child and merge of an assembly, with a
  statistical error guarantee that holds for all candidates at once, so that a planner may safely pick the best one (Sec.~\ref{sec:problem}).
\item (C2) \emph{An analysis of what a test drive reveals:} which predictions are observed and which are extrapolated
  through the motor model, and why the extrapolated share grows with robot size (Sec.~\ref{sec:theory}).
\item (C3) \emph{A method:} a learned estimator of every module's hidden state, a physics simulation that assembles each
  candidate configuration from its modules, and one calibrated safety margin (Sec.~\ref{sec:method}).
\item (C4) \emph{Pre-registered evidence} in simulation (analysis plan and pass/fail bars committed before any
  result was seen): unsafe commitments occurred in \cfConfMainEep{} of episodes against a 10\,\% limit; on lone faulty modules, at each method's own bound, the certificate certified \cfConfCellOursLoneCfown{} against \cfConfCellCapxferLoneCfown{} for the best prior method, with fewer unsafe commitments; \cfConfMergeCf{} of capable never-docked merges were certified; and an uncalibrated physics model failed silently when the motor model was wrong
  (Sec.~\ref{sec:exp}).
\end{itemize}

Such a certificate is the missing input to autonomous reconfiguration: a planner can ask
which configurations are safe to form for each job, receive answers with a stated error rate, and learn which answers rest
on untested assumptions and are worth a short extra test. Nothing in the approach is specific to wheels (Sec.~\ref{sec:limits}); it applies wherever a robot must reason about a
configuration it has not yet formed.
\section{Related work}\label{sec:related}
Modular self-reconfigurable robots, from M-TRAN to Roombots~\citep{murata2002mtran,sprowitz2014roombots}, were
built so that one set of units can take many configurations~\citep{yim2007modular}. Work on these systems has focused on hardware, docking and reconfiguration planning. We supply an input such a
planner needs once modules differ: a certified prediction of what each configuration it could form can do.

Robots that model themselves can detect damage and keep working~\citep{bongard2006resilient,kwiatkowski2019taskagnostic};
\citet{cully2015robots} adapt after damage without needing to diagnose it, and \citet{hu2024reconfigurable} infer
a reconfigurable robot's configuration from its motion. Closest to us is CapXfer~\citep{capxfer2026}, which certifies the capability of a
lone module after a wheel fault from test drives of its assembly. We extend that question from one lone module to every
child and merge an assembly could form, including configurations that have never existed, and to the choice of whether to
test first. 

Morphology-conditioned learning transfers \emph{control} across configurations, through shared modular policies,
self-assembling agents, or transformers over limbs~\citep{huang2020onepolicy,pathak2019selfassembling,gupta2022metamorph,xiong2023modumorph}.
We transfer a calibrated \emph{belief about capability} instead, and our estimator is a set model over wheels~\citep{zaheer2017deepsets}.

Fault diagnosis and fault-tolerant allocation~\citep{blanke2016diagnosis,cristofaro2014fault}
and auxiliary-signal design~\citep{campbell2004auxiliary} treat diagnosis and reconfigurability of a given system;
they do not certify what a \emph{different} configuration will be able to do, at operating points the current configuration may not
reach.

Conformal prediction gives finite-sample, distribution-free coverage under
exchangeability~\citep{vovk2005algorithmic,angelopoulos2023conformal}, and conformal risk control bounds the expected
value of a monotone loss~\citep{angelopoulos2024conformal}. Conformal methods control sequential decision
risk~\citep{lekeufack2024conformal} and make robot planners ask for help or stay
safe~\citep{ren2023robotsask,lindemann2023safeplanning}. \citet{chenni2026domainaware} show that coverage calibrated in simulation can fail silently under shift, and
\citet{schubert2024conformal} predict one vehicle's performance under actuator degradation. We compose calibrated parts through physics, in the spirit
of \citet{yang2023object}.

Amortized simulation-based inference~\citep{cranmer2020frontier,papamakarios2016fast,ramos2019bayessim} must
contend with misspecification~\citep{wehenkel2025addressing}; we measure its effect on a downstream certificate.

Experimental design for decisions targets predictions or downstream utility rather than
parameters~\citep{bickfordsmith2023epig,foster2021dad,huang2024tndp,wagenmaker2021task}. We choose an extra test by its expected value for the commitment decision, over a library of physical tests with time costs.
\section{The robot and the question}\label{sec:problem}
This section fixes the vocabulary used in the rest of the paper: what is hidden, what the robot observes, what
\emph{capability} means, and what the guarantee promises.

\begin{table}[!t]\centering\small
\caption{The Module (Measured Hardware Parameters Used by the Simulators)}\label{tab:module}
\begin{tabular}{@{}ll@{}}\toprule
Module & 24\,cm square, 12.8\,kg, docks on four edges\\
Wheels & 4 mecanum, 100\,mm, 9 rollers at $45^\circ$\\
Motors & 102:1 gear; 3.33\,N\,m stall, 100\,rpm at 12\,V\\
Control & 50\,Hz speed loop per wheel; 11.1\,V battery\\
Sensing & wheel encoders, commanded duty, IMU\\
Test drive & 12\,s: forward/back, sideways, turn\\
Capability & 3-s run per direction at a fixed command\\
\bottomrule\end{tabular}
\end{table}

Each module $k$ has a hidden state $\theta_k$ that the robot cannot read directly. For each of
its wheels the state says whether the wheel is healthy, \emph{weak} (its motor receives only a fraction $g<1$ of the
commanded voltage, as with a degraded driver or connection, so its stall torque and no-load speed both scale by $g$ and its torque--speed line shifts down),
\emph{coasting} (motor leads open: the wheel is dragged against the gearbox's 0.3--1.2\,N\,m back-drive torque) or \emph{braked} (motor terminals shorted: dynamic braking whose torque grows in proportion to wheel speed, on top of the back-drive torque). The state also
includes the small differences every real module has: each motor's stall torque and no-load speed, which vary by up to
$\pm8\,\%$ per wheel on top of $\pm4\,\%$ per module, the module's floor friction, and its battery voltage. The current
configuration, whose modules are docked now, is the \emph{parent}; its arrangement of modules is $m$.

To learn about its modules, the robot runs a short \emph{test drive}: for 12\,s, the whole assembly drives
forward and back, sideways both ways, and turns both ways. It records each wheel's speed and the duty its controller
applied, plus the assembly's motion from one inertial sensor. We write the recordings as $y$; they depend on the hidden
states $\theta$, the arrangement $m$, and sensor noise.

A configuration $B$ is any connected group of modules. Its \emph{capability} in a motion direction $l$, one of
$K=10$ (forward, back, left, right, four diagonals, and turning either way), measures how well it keeps up with a
fixed test command in that direction (0.30\,m/s forward or sideways, 0.21\,m/s along each axis on a diagonal, and a turn
rate set by the configuration's size), carrying its payload if any. Precisely, over a 3-s run with the commanded body velocity $\xi_l=(v_x,v_y,\omega_z)$ (forward speed, sideways speed, turn rate), let $\bar\xi$ be the measured body velocity averaged after the start-up transient. The achieved fraction is $\langle\bar\xi,\xi_l\rangle_W/\|\xi_l\|_W^2$, where $\langle x,z\rangle_W=x_1z_1+x_2z_2+L_c^2x_3z_3$, $\|x\|_W^2=\langle x,x\rangle_W$, and $L_c$ is the largest distance from a wheel to the configuration's centre, so a turn rate counts as the speed it gives the outermost wheel. Capability normalises this by the
same quantity for a nominal configuration $B^{\mathrm{nom}}$ of the same shape (no faults, no motor variation, no
payload):
\begin{equation}\label{eq:rho}
\rho_l(B)=\frac{\langle\bar\xi(B),\xi_l\rangle_W}{\langle\bar\xi(B^{\mathrm{nom}}),\xi_l\rangle_W}.
\end{equation}
So $\rho=1$ means ``as good as a nominal configuration'' and $\rho=0.6$ means the configuration achieves 60\,\% of
what the nominal one does; a payload lowers $\rho$, and $\rho$ can slightly exceed 1. This $\rho_l(B)$ is the quantity every bound must stay below; its true value is known only in simulation, where it labels the calibration episodes and scores the results. A \emph{job} needs $\rho_l(B)\ge\tau$ in one direction; in the experiments $\tau\in\{0.6,0.7,0.8\}$. A job that needs several directions is certified only if each of them is.

The configurations the robot must judge, its \emph{candidates}, are every \emph{child} the parent can split into (any
connected subset of its modules, down to a single module) and every \emph{merge} with another team's assembly,
including assemblies never docked to this one. One \emph{decision} is a (candidate, direction, requirement) triple:
may this configuration be sent on a job that needs $\rho_l(B)\ge\tau$? A certificate answers every decision in the
same way. It computes a \emph{lower bound} $\ell_{B,l}$, its claimed minimum for the true capability, and it
\emph{certifies} the configuration for the job if and only if the bound clears the requirement, $\ell_{B,l}\ge\tau$; otherwise it refuses.

Refusals come in two kinds, and only one of them can be settled by more data: either the estimate $\hat\rho_{B,l}$ (the certificate's best guess before its safety margin, Sec.~\ref{sec:method}) itself falls short of $\tau$, or the estimate clears $\tau$ but the bound, after its uncertainty margin, does not. In the second case a further test could settle the decision, and a robot may choose to run one before deciding; we call this verdict \emph{test first}. It is an exploratory extension (Sec.~\ref{sec:extra}); in the main results every test-first decision counts as a refusal.

\begin{table}[t]\centering\small
\caption{The four outcomes of one decision, illustrated with a requirement $\tau=0.7$. Only the top-right cell is
dangerous. Numbers are illustrative.}\label{tab:outcomes}
\setlength{\tabcolsep}{3pt}
\begin{tabular}{@{}l>{\raggedright\arraybackslash}p{0.34\columnwidth}>{\raggedright\arraybackslash}p{0.34\columnwidth}@{}}\toprule
& Truly capable ($\rho\ge\tau$), e.g.\ $\rho=0.85$ & Truly incapable ($\rho<\tau$), e.g.\ $\rho=0.65$\\\midrule
Certified ($\ell\ge\tau$) & useful: bound $0.75$, job done & \textbf{unsafe commitment}: bound $0.72$ is too high; the
configuration falls short\\
Refused ($\ell<\tau$) & missed opportunity: bound $0.66$, safe but not useful & correct refusal: bound $0.60$\\
\bottomrule\end{tabular}
\end{table}

Table~\ref{tab:outcomes} shows the four possible outcomes. Two quantities describe a certificate: how often it
certifies capable configurations, and how rarely it certifies incapable ones. We report both:
\begin{itemize}
\item the \emph{certified share} (CF): among capable decisions, the fraction certified (higher is more useful);
\item the \emph{unsafe rate} $E_{\mathrm{dec}}$: among incapable decisions, the fraction certified, i.e.\ unsafe
  commitments (lower is safer).
\end{itemize}

The guarantee is stated per \emph{episode}: one test drive of one parent together with all of its candidates (about
20 configurations on average, each in 10 directions; the six-module parent of Fig.~\ref{fig:teaser} has \cfTeaserBodies). A bound
is \emph{too high} if it exceeds the true capability, $\ell_{B,l}>\rho_l(B)$, whether or not any job relies on it. The certificate is built so that, for a risk level $\alpha$ fixed in advance, all bounds in an episode are correct with high probability:
\begin{equation}\label{eq:guarantee}
\Pr\big(\text{any bound in the episode is too high}\big)\le\alpha .
\end{equation}
An unsafe commitment needs a bound that is too high \emph{and} clears the requirement (Table~\ref{tab:outcomes}, top
right), so Eq.~\eqref{eq:guarantee} also limits the quantity that matters for safety, the share of episodes with any
unsafe commitment, which we call $E_{\mathrm{unsafe}}$. The episode-level statement is strict: one unsafe commitment
anywhere among an episode's hundreds of decisions counts the whole episode as unsafe. Because it covers all candidates
at once, the robot may pick the best-looking configuration after seeing the bounds without losing the guarantee. 

A per-bound guarantee, as in standard conformal prediction or in single-module certificates~\citep{capxfer2026}, would
not suffice: with hundreds of decisions per episode, bounds that are each right 90\,\% of the time are almost surely
wrong somewhere, and picking the best-looking configuration would favour exactly the bounds that are too high. We
use $\alpha=0.10$. The probability is over randomly generated parents, faults and variations, the sensor noise, and
the calibration data. We report results at three levels, from strictest to most practical: the share of
episodes with any bound too high (the event of Eq.~\eqref{eq:guarantee}), the share of episodes with any unsafe
commitment ($E_{\mathrm{unsafe}}$), and the per-decision rates CF and $E_{\mathrm{dec}}$.

Before building bounds that meet Eq.~\eqref{eq:guarantee}, Sec.~\ref{sec:theory} asks which of them the test drive can support with data at all.
\section{What a test drive reveals about the next configuration}\label{sec:theory}
\begin{figure}[!t]\centering
\includegraphics[width=\columnwidth]{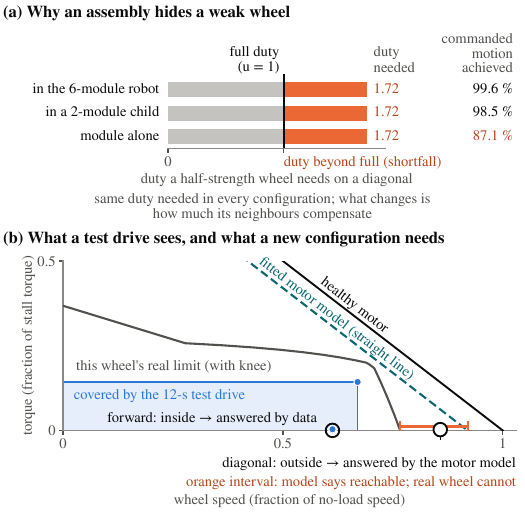}
\caption{Two views of a single wheel (each panel uses a different example wheel), computed in T0. (a) Why an assembly hides a weak wheel. (b) What a test drive
sees and what a new configuration needs; the orange interval marks speeds the fitted model allows but the real wheel cannot reach (from \cfFigconceptSpeedTruetop{} to \cfFigconceptSpeedModeltop{} of no-load speed; the diagonal needs \cfFigconceptSpeedDiag). Both panels are explained in Sec.~\ref{sec:theory}.}\label{fig:concept}
\end{figure}

The answer turns on a single picture: the operating points of one wheel (Fig.~\ref{fig:concept}). To draw it we use an idealised model of an assembly (T0, a \emph{quasi-static} model: steady-state force and torque balance, inertia neglected): every driven wheel either tracks its speed reference while $|u|\le1$ or saturates at $u=\pm1$, and coasting or braked wheels are passive; the configuration then moves at
the speed where the wheel forces balance. In this model the motor model is the standard straight torque--speed line, and one wheel's hidden state reduces to two numbers. Real motors can depart from this line: a current-limited driver or a saturating
motor gives a \emph{knee}, beyond which the effective stall torque and no-load speed fall off linearly with speed or
torque. We call this the \emph{knee motor model}; its severity $s\in\{0,1,2,3\}$ measures how far it departs from the
straight line ($s=0$ is the straight line), and we use it to test what happens when the straight line is wrong.

Figure~\ref{fig:concept}(a) shows why the question arises at all. It follows one wheel at half voltage
($g=\cfFigconceptMaskG$) on a diagonal move: in every configuration its controller would need duty
$u=\cfFigconceptMaskUneedParent>1$, so the wheel saturates and falls short of its share. In the six-module robot its
healthy neighbours drag it along and the configuration still achieves \cfFigconceptMaskRhoParent{} of its commanded
motion; in a two-module child, \cfFigconceptMaskRhoPair; alone, only \cfFigconceptMaskRhoAlone. The same fault is
nearly invisible in the assembly and decisive after the split.

Figure~\ref{fig:concept}(b) shows what the test drive can and cannot tell about that. It plots one wheel's operating
points: wheel speed against the torque the wheel must deliver. The blue region holds every point the 12-s test drive
covered; it ran this wheel up to \cfFigconceptSpeedProbetop{} of its no-load speed, at low torque. A child driving
forward needs \cfFigconceptSpeedForward{} of no-load speed from this wheel, inside the region, so its answer comes
from data. The same child driving diagonally needs \cfFigconceptSpeedDiag, outside the region, so nothing measured
settles it, and the answer comes from the motor model. The straight line fitted to the test drive says the wheel can
reach \cfFigconceptSpeedModeltop{} of no-load speed, which would be enough; but this wheel's real limit has a knee and
tops out at \cfFigconceptSpeedTruetop{} (orange interval). The robot would certify the diagonal with confidence, and be
wrong.

\begin{lemma}[two numbers per wheel]\label{lem:two}
In T0, a driven wheel $i$, healthy or weak and with any motor variation, enters every assembly's equations only
through two unitless constants: $\kappa_i$, the nominal stall torque divided by this wheel's effective stall torque at full duty, and $\nu_i$, the same ratio for no-load speed (both include motor variation, $g_i$ and battery voltage). To deliver torque $a$ (as a fraction of the nominal stall torque) at speed $b$ (as a fraction of the nominal no-load speed), the wheel's controller must apply the duty
\begin{equation}\label{eq:duty}
u_i=a\,\kappa_i+b\,\nu_i,
\end{equation}
and the wheel keeps its commanded speed if and only if $|u_i|\le1$.
\end{lemma}
Lemma~\ref{lem:two} has two consequences. A weak wheel and a slightly slow healthy motor are the same kind of object:
both only change $\kappa_i$ and $\nu_i$. And during the test drive a healthy assembly barely loads its wheels, so
$a\approx0$: the test drive measures $\nu_i$ well and says little about $\kappa_i$. This is why the test drive alone cannot vouch for moves that load a wheel harder, or spin it faster, than the drive did.

To make ``observed'' precise, call an operating point that a new configuration needs from a wheel \emph{dominated} if the test drive
saw that wheel tracking ($|u|\le1$) at a point with at least as much speed and torque; a wheel that tracked there can track here.

\begin{proposition}[what the test drive covers]\label{prop:cover}
(a) Let $E_i(b)$ be the largest torque wheel $i$ can deliver at speed $b$. Suppose we know only that $E_i$ falls with speed and never exceeds a healthy wheel's,
and that a configuration never gets slower when a wheel gets stronger. If every point a configuration needs in a direction is
dominated, the test drive settles whether the configuration meets a requirement. If some point is not dominated and the capabilities of all wheel states that reproduce the recordings span both sides of the requirement, data alone cannot settle it. (Dominance is
sufficient, not necessary: a wheel seen at full duty also pins down its limit.)
(b) For mecanum configurations, moving forward, back or sideways needs exactly the wheel speeds the test drive used, so these
directions are always observed. Diagonal moves drive only half the wheels, each at $1.4\times$ the test speed, so they
are not observed
(on \cfSaDiagUndom{} of children's diagonal moves) unless a higher-speed test drive (\emph{clamp}: 8.8\,s at 0.45\,m/s
forward and sideways) is added. Turning on the spot needs faster outer wheels in children of three or more modules,
and these turns remain unobserved even after clamp (on \cfFigTwoUndomYawMultiClamp{} of them).
(c) In T0, if the test drive sees a wheel at two operating points with linearly independent (torque, speed), both of its constants are
known; if it sees only speed, a needed point is settled only if the answer is the same for every $\kappa_i$ in its physically possible range.
\end{proposition}

\begin{figure}[t]\centering
\includegraphics[width=\columnwidth]{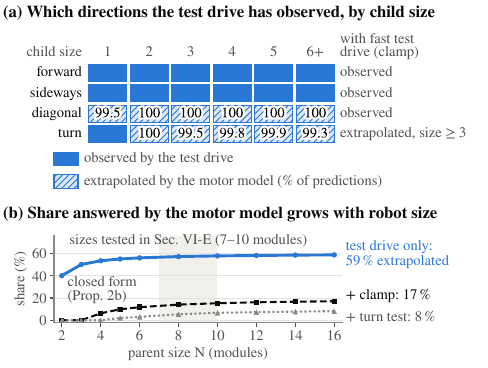}
\caption{Where the answers come from. (a) For each direction and child size, the share of predictions the standard test
drive has observed (solid) or that are extrapolated by the motor model (hatched). With the standard test drive, turns of every child of two or more modules are extrapolated; adding the clamp test covers pairs but not children of three or more (``with fast test drive'' column). (b) The extrapolated
share grows with the size of the parent, for the standard drive, with clamp added, and with clamp plus a turn at twice
the standard rate (``+ turn test''). Closed-form computation in T0 (Prop.~\ref{prop:cover}b).}
\label{fig:provenance}
\end{figure}

Proposition~\ref{prop:cover} sorts every prediction about a new configuration into two kinds. \emph{Observed} predictions are
settled by data the test drive collected. \emph{Extrapolated} predictions are settled by the motor model, and a robot
that takes its motor model at face value will give them as confidently as observed ones. Figure~\ref{fig:provenance}
shows where each kind lives. Because a large parent turns slowly while its small children turn fast, the extrapolated
share grows with the parent's size: with the standard test drive it rises from \cfFigTwoShareVtwoNFour{} for four
modules to \cfFigTwoShareVtwoNSixteen{} for sixteen, and adding clamp leaves \cfFigTwoShareClampNSixteen{}, all of it on
turns. This predicts where an uncalibrated physics model will fail first, and silently, when the motor model is wrong (Sec.~\ref{sec:gate}). A certificate must therefore stay valid where its model is wrong; Sec.~\ref{sec:method} builds one by calibrating the whole pipeline rather than trusting any part.

\section{Method: from one test drive to a verdict for every configuration}\label{sec:method}
\begin{figure*}[!t]\centering
\includegraphics[width=\textwidth]{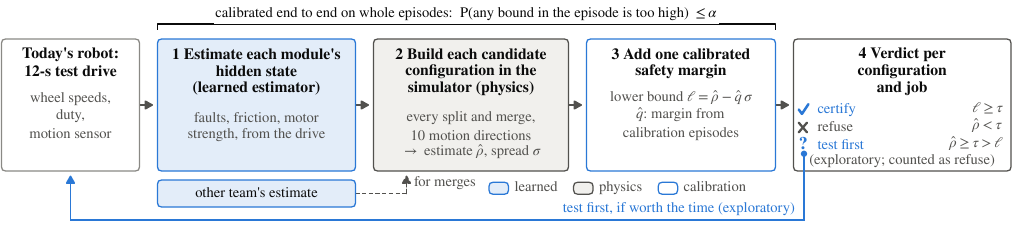}
\caption{From one test drive to a verdict for every configuration. (1) A learned estimator reads the recordings and outputs a
probability distribution over each module's hidden state. (2) A physics simulator assembles every child and merge from
those modules (merges also use the other team's estimate) and predicts each configuration's capability in every direction. (3) One
safety margin, calibrated on whole episodes, turns the predictions into lower bounds that are correct together with
probability at least $1-\alpha$. (4) Each configuration gets a verdict per job. Where a bound straddles the requirement, the
robot may run a further test if it is worth the time (exploratory).}\label{fig:pipeline}
\end{figure*}
The certificate works in three steps (Fig.~\ref{fig:pipeline}).

\emph{Step 1: a learned estimator of each module's state.} We train a neural network once, in simulation, to map any test drive of any assembly directly to a probability distribution over its modules' states (an \emph{amortized} posterior), which avoids running the simulator thousands of times per decision, a cost of minutes per episode. The network treats the assembly as a set of wheels. Each wheel's recorded speed and duty traces are
summarised by a small convolutional encoder and tagged with the wheel's position, and a transformer lets the wheels
exchange information through self-attention. Because it sees positions rather than a fixed wheel numbering, one network serves every shape. For each
wheel it outputs the probability of each condition (healthy, weak, coasting, braked) and a distribution over the
weak wheel's voltage fraction $g$; for each module, distributions over friction and motor variation. We train it on
simulated test drives of parents with up to six modules, randomising every physical parameter widely.

\emph{Step 2: assemble each candidate in simulation.} Composition is physics, not learning. For a candidate configuration $B$
we draw 16 samples of its modules' fault modes, $g$ and module parameters from the estimator, and of per-wheel motor variation and battery voltage from the simulator's prior, build the configuration in our simulator T1 (a fast dynamic simulator identified from real-fleet logs,
Sec.~\ref{sec:setup})
and drive it on all ten directions. The median of the 16 predicted capabilities is the prediction $\hat\rho_{B,l}$, and the
gap down to their 5th percentile is its spread $\sigma_{B,l}$. Because configurations are assembled from modules, the same
procedure serves children, merges and shapes never seen in training.

\emph{Step 3: one calibrated safety margin.} The predictions are not yet trustworthy: the estimator is approximate and
the simulator imperfect. We correct both at once with \emph{conformal calibration}~\citep{vovk2005algorithmic}. On each
of $n$ calibration episodes $e$, where the true capabilities are known, we record the worst over-prediction across all
of that episode's configurations and directions, in units of spread (a positive $r_e$ means some prediction in episode $e$ was too optimistic):
\begin{equation}\label{eq:score}
r_e=\max_{\text{configurations }B,\ \text{directions }l\text{ of episode }e}\ \frac{\hat\rho_{B,l}-\rho_l(B)}{\sigma_{B,l}} .
\end{equation}
The margin $\hat q$ is the value that $1-\alpha$ of these worst cases stay below (precisely, the $k$-th smallest of the $n$ scores, $k=\lceil(n+1)(1-\alpha)\rceil$, where $\lceil\cdot\rceil$ rounds up; for $n=100$ and $\alpha=0.1$ this is the 10th largest), and each bound is
\begin{equation}\label{eq:lower}
\ell_{B,l}=\hat\rho_{B,l}-\hat q\,\sigma_{B,l}.
\end{equation}
The configuration is certified for a job in direction $l$ only if $\ell_{B,l}\ge\tau$.

\begin{proposition}[the guarantee]\label{prop:valid}
If the calibration episodes and a new episode are exchangeable (their joint distribution does not change if they are reordered, as for independent draws), the bounds of
Eq.~\eqref{eq:lower} satisfy Eq.~\eqref{eq:guarantee}, for all configurations, directions and requirements of the episode at once,
and for any rule that picks among the configurations after seeing the bounds.
\end{proposition}
The argument is standard~\citep{lindemann2023safeplanning}: a new episode's worst case is equally likely to fall
anywhere among the calibration worst cases, so it exceeds $\hat q$ with probability at most $\alpha$, and if it does
not, every bound in the episode is correct (Appendix~\ref{app:proofs}).  

Merges add one more source of error to this pipeline. The state of each team's modules is estimated from that team's own test drive,
before anyone decides which merge to form, so those estimates serve every merge at once. What remains specific to a
merge is the \emph{composition error}: how far a merged configuration differs from the configuration simulated from its modules' states because of flexible
docks and shifted loads. Theorem~\ref{thm:merge} (Appendix~\ref{app:proofs}) shows that a merge bound then fails with probability at most the teams' calibration errors plus a shape margin and this composition error; the merge margin we evaluate is fitted empirically rather than built from those terms. We measure it directly in Sec.~\ref{sec:merge}.

In our merge study we enumerate each team's fault hypotheses exactly, assemble the merged
configuration in T1, and lower the result by a margin with two parts, a conformal term and a term per merge size, both fitted on earlier exploratory merges so that merges are scored at the matched operating point
(Sec.~\ref{sec:setup}).

An exploratory extension, evaluated only in T0, decides when to test first using the \emph{expected value of information} (EVOI) of the test for the decision: when a bound straddles a requirement, the robot can run the faster
clamp test (8.8\,s) first, but only when the expected gain in correctly certified jobs, valued in seconds of work,
exceeds the test's time. Section~\ref{sec:exp} tests each of these pieces in turn.
\section{Experiments}\label{sec:exp}
In one pre-registered run, the guarantee held (\cfConfMainEep{} of episodes with an unsafe commitment, against a 10\,\% limit), the certificate beat the best prior method on lone faulty modules, and it certified never-docked merges. Sections~\ref{sec:cert}--\ref{sec:extra} take these and the remaining questions in turn, and Table~\ref{tab:glance} collects the answers.
\subsection{Setup}\label{sec:setup}
All experiments use three simulators of the same module (Table~\ref{tab:module}): a 24\,cm square platform with four
mecanum wheels in an X layout, a speed controller on each wheel, and edge docks. T0 is the idealised quasi-static model of
Sec.~\ref{sec:theory}, used for analysis. T1 is a fast dynamic simulator whose parameters were identified from logs of
a real fleet of six such modules, extended with per-wheel and per-module motor variation, friction, battery voltage,
the random split of normal load among wheels, and payload; it is used for training, calibration and testing. T2 is an independently written
simulator with different numerics and effects T1 lacks, such as flexible docks with play, load shifting, and rollers
whose drag depends on direction. Optionally, its motors follow the knee motor model of Sec.~\ref{sec:theory}. T2 is never used to train our estimator or to calibrate our
certificate; the physics certificates of Sec.~\ref{sec:gate} were calibrated on separate exploratory T2 episodes.

An episode draws a parent of 1--6 modules from 37 shapes in five classes (chains, L-shapes, paths, loops, grids), up to
three faulty wheels (weak with voltage fraction $g$ between 0.3 and 0.95, coasting, or braked), motor variation (up to
$\pm8\,\%$ per wheel on top of $\pm4\,\%$ per module) and the other variations of Sec.~\ref{sec:problem}, and one or two other teams to merge with. Half of the faulty wheels are weak,
a quarter coast and a quarter are braked, and a parent has none, one, two or three faulty wheels with probabilities
0.2, 0.4, 0.3 and 0.1; the coasting wheel in Fig.~\ref{fig:teaser} is one example among these. The draws are
deliberately rich in faulty wheels so that hard cases are well represented; the
guarantee holds for this mix, and a fleet with rarer faults would be calibrated on its own episodes. An episode has
about 20 candidate configurations, each judged on ten directions at three requirements. To test beyond training, we also use parents
of 7--10 modules and two shifts in wheel type: parents whose rollers grip five times less along the roller axis than in training, and single parents that mix mecanum and conventional modules. For the wheel-type tests, the
estimator is a version trained on single-type assemblies with each wheel's type as an input.

We compare against methods that span trivial to strong. \emph{CapXfer}~\citep{capxfer2026} is the strongest prior
certificate: it bounds each wheel's state with a learned fault model and composes those per-wheel bounds into the
child; we recalibrate it with the same margin as ours. Simpler rules: \emph{$\hat g$-physics} fits each wheel's voltage
fraction by least squares and predicts with T0; \emph{min over members} takes, per direction, the weakest module on its
own; \emph{copy parent} reuses the parent's measured capability; \emph{detector-gated} certifies only configurations with no
detected fault; \emph{shape metadata} predicts from size and direction alone; an \emph{oracle} sees the true state.

The main claims were tested in a pre-registered run: the analysis plan, with $\alpha=0.10$, $U^\ast=1\,\%$, every
pass/fail bar and a multiple-testing correction (Holm, over seven main comparisons, P1--P7), was committed before any test
result was seen, and the run was scored once by the committed program. Confidence intervals resample whole episodes.
Our method ran on \cfConfMainNeps{} test episodes and the baselines on \cfConfCellCapxferNeps; results outside this run are labelled exploratory. Amendments and disclosures are in Appendix~\ref{app:prereg}.

Most candidates are healthy and far above every requirement, which makes any method look good, so we compute per-decision rates on the \emph{hard cases}: configurations that contain a faulty wheel, or whose capability lies
within $0.1$ (in units of $\rho$) of the requirement. These cases are defined from the true capability, so they are used only to score
methods, never to make decisions. Methods are compared at their own guaranteed bound, and also at a \emph{matched point}, where each method's margin is retuned on calibration data so that its unsafe rate should be
$U^\ast=1\,\%$; this compares usefulness at a common target risk. For shifted test sets we also report \emph{coverage}, the share of hard-case bounds that are correct.
Throughout, CF is a share of capable decisions, $E_{\mathrm{dec}}$ a share of incapable decisions (both on the hard
cases), and $E_{\mathrm{unsafe}}$ a share of episodes.

\subsection{Does the guarantee hold, and is the certificate useful?}\label{sec:cert}

The safety guarantee held, and at its own bound our certificate was both more useful and safer than CapXfer. Only \cfConfMainEep{} of the \cfConfMainNeps{} test episodes contained any unsafe commitment, against the pre-registered limit of 10\,\% (Table~\ref{tab:main}).
\begin{table*}[!t]\centering\small
\caption{Main Test: Certified Share and Unsafe Rates, at Each Method's Own Bound and at the Matched Point}
\label{tab:main}\begin{tabular}{@{}lrrrrrrrrrr@{}}\toprule
& & \multicolumn{5}{c}{Own bound ($\alpha=0.10$)} & \multicolumn{4}{c}{Matched point (target $U^\ast=1\,\%$)}\\\cmidrule(lr){3-7}\cmidrule(l){8-11}
& & \multicolumn{3}{c}{Hard cases} & \multicolumn{2}{c}{Faulty module alone} & \multicolumn{2}{c}{Hard cases} & \multicolumn{2}{c}{Faulty module alone}\\\cmidrule(lr){3-5}\cmidrule(lr){6-7}\cmidrule(lr){8-9}\cmidrule(l){10-11}
Method & Episodes & CF & $E_{\mathrm{dec}}$ & $E_{\mathrm{unsafe}}$ & CF & $E_{\mathrm{dec}}$ & CF & $E_{\mathrm{dec}}$ & CF & $E_{\mathrm{dec}}$\\\midrule
\textbf{Ours} & 1,050 & \textbf{95.3} & 0.21 & 0.95 & \textbf{72.8} & 0.12 & 98.0 & 3.21 & 85.2 & 2.89 \\
\midrule
CapXfer & 210 & 84.7 & 3.22 & 2.86 & 51.8 & 0.99 & 82.4 & 2.39 & 46.7 & 0.77 \\
$\hat g$-physics & 210 & 64.1 & 2.03 & 6.19 & 50.3 & 1.62 & 62.8 & 1.43 & 49.7 & 1.34 \\
Min over members & 210 & 36.3 & 1.43 & 4.76 & 50.3 & 1.62 & 35.2 & 1.19 & 49.9 & 1.34 \\
Copy parent & 210 & 0.0 & 0.00 & 0.00 & 0.0 & 0.00 & 8.8 & 1.25 & 11.2 & 1.48 \\
Detector-gated & 210 & 9.0 & 0.83 & 0.95 & 20.9 & 0.99 & 5.4 & 0.12 & 14.1 & 0.14 \\
Shape metadata & 1,050 & 0.0 & 0.00 & 0.00 & 0.0 & 0.00 & 0.0 & 0.00 & 0.0 & 0.00 \\
Oracle (full state) & 210 & 100.0 & 0.00 & 0.00 & 100.0 & 0.00 & 100.0 & 0.66 & 100.0 & 0.42 \\
\bottomrule\end{tabular}

\vspace{2pt}\begin{minipage}{0.96\textwidth}\footnotesize Pre-registered run on simulator T1, in \%. CF: certified share
of capable decisions; $E_{\mathrm{dec}}$: unsafe rate among incapable decisions; $E_{\mathrm{unsafe}}$: share of
episodes with any unsafe commitment. Left: each method at its own bound ($\alpha=0.10$). Right: each method's margin
retuned on calibration data for $U^\ast=1\,\%$.
CapXfer and most baselines ran on \cfConfCellCapxferNeps{} episodes, \cfConfPosthocNcommon{} shared with ours.\end{minipage}
\end{table*}
\begin{figure*}[!t]\centering
\includegraphics[width=\textwidth]{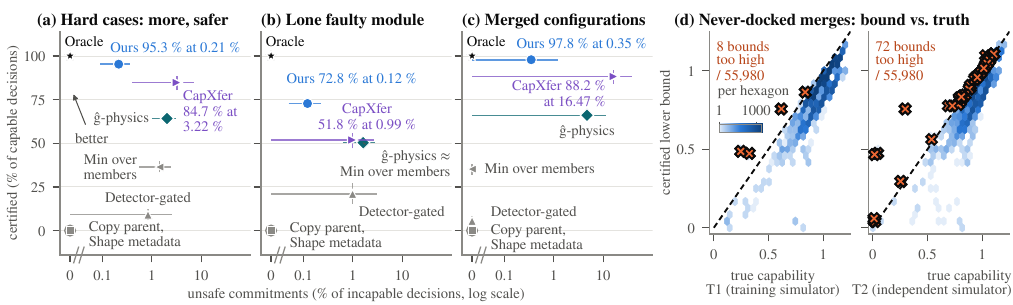}
\caption{Our certificate certifies more capable configurations and fewer incapable ones. Each point is a method at its
own bound: certified share against unsafe rate, with 95\,\% intervals; up and to the left is better. (a) Hard cases;
(b) faulty modules left alone after a split, the case CapXfer was designed for; (c) merged configurations in the main test.
Unsafe rates are omitted where fewer than 50 incapable decisions exist. (d) Merges of teams never docked together:
certified lower bound against true capability for every (merge, direction); points above the diagonal are bounds that were
too high (\cfFigFourMergeMissesTOne{} on T1, \cfFigFourMergeMissesTTwo{} on T2).}\label{fig:ownbound}
\end{figure*}

On the hard cases, ours certified \cfConfCellOursStratumCfown{} of capable decisions against CapXfer's
\cfConfCellCapxferStratumCfown, while certifying only \cfConfCellOursStratumEdecown{} of incapable ones against
\cfConfCellCapxferStratumEdecown. On faulty modules left alone after a split, the case CapXfer was designed for, the
gap widens: \cfConfCellOursLoneCfown{} against \cfConfCellCapxferLoneCfown{} certified, at \cfConfCellOursLoneEdecown{}
against \cfConfCellCapxferLoneEdecown{} unsafe. Ours is better on both axes at once (Fig.~\ref{fig:ownbound}), and the
pre-registered comparisons, made at the matched point (Table~\ref{tab:main}, right), point the same way: \cfConfMainPOne{} points more lone faulty modules certified than
CapXfer, and \cfConfMainPTwo{} points more hard cases than the best simple rule, $\hat g$-physics (Holm-corrected;
details in Appendix~\ref{app:prereg}).

The simple rules fall short for reasons that show why a certificate is needed:
copy parent and shape metadata cannot see that a module is weak, so to stay safe they certify almost nothing;
detector-gated refuses any configuration with a detected fault, including many that can still do the job; and
$\hat g$-physics attaches the same uncertainty to every prediction, however well its fit is determined.

Every kind of configuration stayed at about 1\,\% unsafe or below (merges \cfConfCellOursMergeEdecown, multi-module
children \cfConfCellOursMultiEdecown, lone modules \cfConfCellOursLoneEdecown). In practice, the robot can commit to
the best certified configuration and rarely be wrong, while giving up far fewer usable configurations than the prior
certificate.

\subsection{Configurations that never existed}\label{sec:merge}
The certificate also certifies configurations that have never existed. In a separate study of merges of teams never docked together, for which no data exist, it certified \cfConfMergeCf{} of capable merge decisions at the matched point (Fig.~\ref{fig:ownbound}c shows merges within the main test instead, at the own bound). Assembling the merged
configuration from its parts loses almost nothing: across \cfConfMergeNpairs{} pairs of teams, building the merge from
the two teams' true module states changed \cfConfMergeFlipTOne{} of decisions on T1 and \cfConfMergeFlipTTwo{} on the
independent T2. That compositional certificate, with fault conditions inferred from each team's own test drive, certified
\cfConfMergePFour{} points more than copying each parent's answer (pre-registered). A bound was too high in
\cfConfMergeMiscovTOne{} of pairs on T1 and \cfConfMergeMiscovTTwo{} on T2, within the pre-registered bars of 10\,\% and
15\,\%. Merges are also where the prior method struggles most: at its own bound in the main test, CapXfer certified
\cfConfCellCapxferMergeEdecown{} of incapable merge decisions, against \cfConfCellOursMergeEdecown{} for ours
(Fig.~\ref{fig:ownbound}c). A team can therefore plan merges it has never tried from its members' test drives alone.

\begin{table*}[!t]\centering\small
\caption{Summary of Results}\label{tab:glance}
\renewcommand{\arraystretch}{1.25}
\begin{tabular}{@{}>{\raggedright\arraybackslash}p{0.32\textwidth}>{\raggedright\arraybackslash}p{0.50\textwidth}>{\raggedright\arraybackslash}p{0.13\textwidth}@{}}\toprule
\textbf{Question} & \textbf{Key result} & \textbf{Outcome}\\\midrule
Are unsafe commitments rare? & \cfConfMainEep{} of episodes had one (bound 10\,\%) & Met\\

Better than CapXfer on lone faulty modules? & \cfConfCellOursLoneCfown{} vs.\ \cfConfCellCapxferLoneCfown{} certified, \cfConfCellOursLoneEdecown{} vs.\ \cfConfCellCapxferLoneEdecown{} unsafe & Better on both\\
Can it certify never-docked merges? & \cfConfMergeCf{} certified; a bound too high in \cfConfMergeMiscovTOne{} of pairs & Met\\

What if the motor model is wrong? & Uncalibrated physics silently certifies \cfConfGateRawUndom{} of incapable extrapolated decisions; with a calibrated margin, \cfConfGateParUndom & As predicted (C2)\\
Larger configurations, new wheels? & Unsafe rates within bars; coverage \cfConfScaleCovNine{} at 9--10 modules & Safe; coverage below bar\\
Is an extra test worth its time? & Same jobs served with \cfProbeExtraEvoi\,s instead of 8.8\,s of testing & Exploratory\\
\bottomrule\end{tabular}

\vspace{3pt}\begin{minipage}{0.94\textwidth}\footnotesize Rows follow Sec.~\ref{sec:cert}--\ref{sec:extra}. All rows except the last come from the single pre-registered run.\end{minipage}
\end{table*}
\subsection{When the motor model is wrong}\label{sec:gate}
\begin{figure}[t]\centering
\includegraphics[width=\columnwidth]{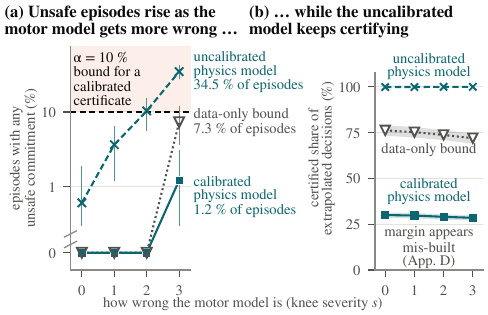}
\caption{An uncalibrated physics model fails silently when its motor model is wrong. Children of three or more
modules on T2, whose motors follow the knee model with increasing severity. (a) Share of episodes with any unsafe
commitment; (b) certified share of capable extrapolated decisions. The uncalibrated model keeps certifying while its share of unsafe episodes climbs; with a calibrated margin it stays safe; a bound that uses only observed points rises sharply at $s=3$. Pre-registered run.}\label{fig:silent}
\end{figure}
Calibration keeps a physics model safe when its motor model is wrong; taken at face value, the same model fails silently, as Sec.~\ref{sec:theory} predicts. At the strongest knee, \cfFigFiveRawSThreeEep{} of episodes contained an unsafe commitment for the uncalibrated model, against \cfFigFiveParSThreeEep{} with a calibrated margin (Fig.~\ref{fig:silent}), and nothing in the uncalibrated model's own output signals the failure. 

To isolate the motor model, we compared three certificates built on the same straight-line physics, fitted to the test drive and told which wheels are faulty: uncalibrated (the prediction is the bound), calibrated (a margin fitted
on straight-line motors), and a data-only bound restricted to operating points observed after the clamp test. They
ran on T2 with knee motors of increasing severity $s$, on children of three or more modules, where
Prop.~\ref{prop:cover}(b) places the extrapolated directions. With straight-line motors ($s=0$) the uncalibrated model
had an unsafe commitment in \cfFigFiveRawSZeroEep{} of episodes; at $s=3$ it certified \cfConfGateRawUndom{} of the
incapable extrapolated decisions while still certifying almost all capable ones (Fig.~\ref{fig:silent}). The same physics with a calibrated margin stayed safe, certifying only \cfConfGateParUndom{} of those decisions.

This is why our margin is calibrated on the output of the whole pipeline rather than on trust in one model: motor errors present in the calibration episodes widen the margin, and the rest surface as extrapolated answers the robot can test first (Sec.~\ref{sec:extra}).

\subsection{Larger configurations and new wheel types}\label{sec:transfer}
Beyond its training distribution the certificate stayed safe. The share of episodes with an unsafe commitment stayed within its pre-registered bars:
\cfConfScaleEepSeven{} and \cfConfScaleEepNine{} for parents of 7--8 and 9--10 modules (training used at most six), and
\cfConfHetEepHone{} and \cfConfHetEepHtwo{} for low-grip rollers and mixed wheel types. Coverage fell from
\cfConfScaleCovSeven{} to \cfConfScaleCovNine{} at 9--10 modules, below its 85\,\% bar, which is the erosion
Sec.~\ref{sec:theory} predicts as the extrapolated share grows; on mixed wheel types the certified share fell by
\cfConfHetCfdropHtwo{} points. 

\subsection{When is an extra test worth its time? (exploratory)}\label{sec:extra}
Running the clamp test only when its expected value for the decision exceeds its time serves almost the same jobs with far less testing (exploratory, T0 only). With uncertain motor models, this rule served \cfTFourUsefulEvoi{} of jobs against
\cfTFourUsefulFixedVTwoClamp{} for always testing, while spending \cfProbeExtraEvoi\,s instead of 8.8\,s of extra testing
per episode on average; the result replicated on three disjoint episode sets (Appendix~\ref{app:explore}). The gain is
time saved, not better decisions.

\subsection{What the results show}\label{sec:show}
Table~\ref{tab:glance} collects the answers. Taken together, the experiments support each contribution of Sec.~\ref{sec:intro} as follows. The problem of C1 can be
solved in practice: a certificate that answers for every child and merge at once, from one 12-s test drive, kept unsafe
commitments to \cfConfMainEep{} of episodes, and its guarantee covers selection, so a planner may pick the best
certified configuration. The analysis of C2 made two predictions, and the data support both: answers that rest on the
motor model fail silently when that model is wrong (Sec.~\ref{sec:gate}), and coverage erodes as configurations grow
and more answers are extrapolated (Sec.~\ref{sec:transfer}). The method of C3 is more useful than the best prior
certificate at lower risk, and it extends to merges of teams never docked together, which no single-module method
addresses.
All of this evidence comes from the single pre-registered run (C4). Section~\ref{sec:limits} asks how far these results reach beyond wheels and beyond simulation.
\section{Generality and limitations}\label{sec:limits}
Nothing in the certificate is specific to wheels or to the faults we simulated. Its three steps (Fig.~\ref{fig:pipeline}) need only per-part hidden states, a simulator that composes parts, and calibration episodes that resemble deployment. The
observed-versus-extrapolated analysis carries over too: for any actuator, a test exercises some operating points, a new
configuration may need others, and those answers lean on a model of the actuator. A legged module would replace wheel
speed and torque by joint speed and torque; an aerial module, by rotor speed and thrust. Other faults, such as sensor
bias or a drifting gearbox, enter as further hidden states to estimate, and other questions, such as whether a merged
team can carry a heavier payload or a new tool, enter as further capabilities to bound. Prop.~\ref{prop:cover}(b) is
specific to mecanum wheels, and none of these extensions is tested here.

All evidence is from simulation, and the extra-test study uses the idealised model T0. The guarantee holds over the
distribution the episodes were drawn from. Its calibration set was small (100 episodes), individual bounds were too high in \cfConfPosthocMiscov{} of episodes (nominal 10\,\%; most such overshoots did not change a decision), and the matched point's 1\,\% target, the data-only bound at the strongest motor error, coverage at 9--10 modules and usefulness on mixed wheel types fell short of their bars, and nothing beyond 10 modules was tested (Appendix~\ref{app:prereg}). A second
pre-declared run with a twenty-fold larger calibration set and all methods on the same episodes is under way and includes our full certificate under knee motors on T2; the next step is to score real splits and merges on the fleet behind T1.

\section{Conclusion}\label{sec:concl}
A docked team need not reconfigure to learn what it could do. From one 12-s test drive, it can certify which of its
possible children and merges can do a job, including configurations that have never existed, at a stated error rate.
It also knows which of those answers rest on data and which on its motor model, and are therefore worth a short test
first. That is the input an autonomous reconfiguration planner has been missing.


\appendices
\section{Proofs}\label{app:proofs}
This appendix supports Secs.~\ref{sec:theory} and~\ref{sec:method}: why one wheel reduces to two constants
(Lemma~\ref{lem:two}), why dominance decides what a test drive settles (Prop.~\ref{prop:cover}), why one margin covers
every configuration of an episode (Prop.~\ref{prop:valid}), and why merges add only a composition error
(Thm.~\ref{thm:merge}).
\paragraph{Lemma~\ref{lem:two}.} In T0 a driven wheel's torque is
$T=T_s k^T_i\,(u\,g_i v_b-\omega/(\omega_0 k^\omega_i))$ (stall torque $T_s k^T_i$ and no-load speed $\omega_0 k^\omega_i$, with $k^T_i,k^\omega_i$ this motor's variation factors, and $v_b$ the battery voltage divided by its nominal value), linear in the duty $u$ and the speed $\omega$. Normalising load
torque $a=T/T_s$ and speed $b=\omega/\omega_0$ and solving for $u$ gives
$u=a/(g_iv_bk^T_i)+b/(g_iv_bk^\omega_i)=a\kappa_i+b\nu_i$. The wheel tracks iff the required duty is feasible, $|u|\le1$.
Every assembly equation involves the wheel only through this relation, so $(\kappa_i,\nu_i)$ is sufficient. \qed

\paragraph{Proposition~\ref{prop:cover}(a).} Observing tracking at $(a_p,b_p)$ reveals only $E_i(b_p)\ge a_p$. The step
$E^{\mathrm{low}}(b)=\max\{a_p:b_p\ge b\}$ and the healthy envelope are both non-increasing and consistent with all
observations, and so is every envelope between them. By the monotonicity assumption (more torque never
lowers $\rho$; in T0 violations are rare, and their rate enters Thm.~\ref{thm:merge} as $\delta_M$), $\rho_l(B)$ lies between its values at the two ends, and both ends are attained by consistent envelopes. If every
needed point is dominated, every consistent envelope lets every wheel track there, so all give the same $\rho$ and the
decision is identified. If the interval straddles the requirement $\tau$, consistent envelopes near the low end make any commitment
unsafe, so the observations alone cannot license it.  \qed

Part (b) is a closed-form computation: for X-mecanum wheels with roller axes at $45^\circ$, the wheel speed is
$|\omega_i|=\sqrt2\,|v_i\cdot e_i|/R$ with $v_i$ the contact velocity, $e_i$ the unit vector along wheel $i$'s roller axis and $R$ the wheel radius. Translations need speeds independent of
position, hence equal to the test drive's; diagonals at the commanded speeds need $1.4\times$ the test drive's wheel speed on half
the wheels; turns at a rate capped by the child's size need speeds that grow with the distance to the child's own
centroid. We verified the closed form against the T0 inverse kinematics on 200 random parents and all their children
(exact to numerical precision). Part (c) follows from Eq.~\eqref{eq:duty}: with noise-free duties, the values of $(\kappa_i,\nu_i)$ consistent with the data lie on a line in the $(\kappa_i,\nu_i)$ plane, within their physically possible range; two independent
points fix it, and with speed points only ($a\approx0$) it fixes $\nu_i$ and leaves $\kappa_i$ in its prior range, so a
needed condition $a_c\kappa_i+b_c\nu_i\le1$ is identified only if it holds, or fails, over that whole range.

\paragraph{Proposition~\ref{prop:valid}.} Let $r_1,\dots,r_n$ be the calibration scores of Eq.~\eqref{eq:score} and
$r_{n+1}$ the score of a test episode. Under exchangeability the rank of $r_{n+1}$ is uniform, so
$\Pr(r_{n+1}>\hat q)\le\alpha$ for $\hat q$ the $k$-th smallest score, $k=\lceil(n+1)(1-\alpha)\rceil$. If
$r_{n+1}\le\hat q$, then $\hat\rho_{B,l}-\rho_l(B)\le\hat q\sigma_{B,l}$, i.e.\ $\ell_{B,l}\le\rho_l(B)$, for every
$(B,l)$ of the episode, so no commitment can be unsafe for any requirement $\tau$ and under any selection. \qed

\begin{theorem}[compositional merge certificate]\label{thm:merge}
Let assemblies $G_1,\dots,G_J$ be tested separately. For each, let $\mathcal{E}_j$ be the event that assembly $j$'s
certificate gives, for every wheel, a torque--speed limit no higher than the true one, failing with probability
$\Pr(\text{not }\mathcal{E}_j)\le\alpha_1^{(j)}$; let a margin that depends only on the configuration's shape turn
these limits into a lower bound $\ell_l(B)$ for any composed configuration $B$, failing with probability at most
$\alpha_2$; let $\delta_M$ bound the probability that capability decreases when a wheel gets stronger, and
$\delta_{\mathrm{comp}}$ the probability that composition from the parts differs from the merged plant. Then for every child or merge $B$,
$\Pr(\rho_l(B)<\ell_l(B))\le\sum_j\alpha_1^{(j)}+\alpha_2+\delta_M+\delta_{\mathrm{comp}}$, and the events
$\mathcal{E}_j$ do not depend on which $B$ is later formed.
\end{theorem}
\noindent\emph{Proof.} When every $\mathcal{E}_j$ holds, every wheel's true torque--speed limit lies above its certified one; by
monotonicity (except with probability $\delta_M$) the composed capability is at least that of the certified limits,
which the shape margin bounds below except with probability $\alpha_2$; composition matches the merged plant except with
probability $\delta_{\mathrm{comp}}$. Adding the failure probabilities gives the result. \qed

\noindent Only the per-assembly stage holds for all merges at once; the shape margin is per configuration. The construction follows the
sequential compositional certificate of \citet{capxfer2026}. The merge certificate we evaluate does not instantiate
these constants: it uses an empirical size-dependent margin fitted on earlier exploratory merges.

\section{Motor model and simulators}\label{app:sim}
This appendix gives the exact forms behind the knee motor model (Sec.~\ref{sec:theory}) and the simulators
(Sec.~\ref{sec:setup}).

\emph{Knee motor model.} Beyond a speed knee $\omega_k=f_w\,\omega_0$ and a torque knee $T_k=f_t\,T_s$, a wheel's
effective no-load speed and stall torque are scaled by $\max\{0.4,\,1-c_w(|\omega|/\omega_k-1)_+\}$ and $\max\{0.4,\,1-c_t(|T|/T_k-1)_+\}$, where $(x)_+=\max\{x,0\}$. Each severity draws $(f_w,c_w,f_t,c_t)$ per wheel from the ranges in
Table~\ref{tab:knee}; $s=0$ is the straight line of Lemma~\ref{lem:two}.
\begin{table}[h]\centering\small
\caption{Knee Motor Model Parameter Ranges}\label{tab:knee}
\begin{tabular}{@{}ccccc@{}}\toprule
$s$ & $f_w$ & $c_w$ & $f_t$ & $c_t$\\\midrule
1 & 0.75--0.95 & 0.3--0.7 & 0.30--0.50 & 0.3--0.7\\
2 & 0.60--0.90 & 0.6--1.2 & 0.20--0.40 & 0.6--1.2\\
3 & 0.50--0.80 & 1.0--2.0 & 0.10--0.30 & 1.0--2.0\\\bottomrule
\end{tabular}
\end{table}

\emph{T1 and T2.} T1 integrates the wheel--floor contact with a tanh slip law, roller drag, the PI speed loop and
actuation latency at a 5-ms step; its parameters were identified from logs of the real fleet. T2 was written
separately and adds what T1 lacks: dock compliance per face (axial $2\times10^4$--$2\times10^5$\,N/m, lateral
$5\times10^3$--$5\times10^4$\,N/m, yaw 50--1000\,N\,m/rad) with 1.1\,mm hook play; load transfer from a centre of mass
0.04--0.11\,m high; and direction-dependent roller drag (scaled by up to $4\times$ either way). These are the
effects that make a merged configuration differ from the one composed from its modules (Sec.~\ref{sec:merge}).

\section{Estimator and calibration details}\label{app:est}
This appendix expands Sec.~\ref{sec:method}.

\emph{Estimator.} Each wheel token is a temporal-convolution encoding of that wheel's speed reference, commanded duty
and measured speed, plus an encoding of its position; one assembly token encodes the body commands and inertial
signal. Three self-attention layers (width 96, four heads) mix the tokens. Heads output per-wheel fault-mode
probabilities and a Beta distribution over $g$, and per-module Gaussians over friction and motor scales. Training
minimises the fault-mode cross-entropy and the Beta and Gaussian negative log-likelihoods, on test drives simulated on the fly from parents of up to six modules in 37 shape families.

\emph{Calibration, worked through.} With $n=100$ calibration episodes and $\alpha=0.10$, the margin $\hat q$ is the
$\lceil 101\times0.9\rceil=91$st smallest worst-case score, that is, the 10th largest. A new episode's worst case is
equally likely to rank anywhere among the 101, so it exceeds $\hat q$ with probability at most $10/101<\alpha$. This
is an average over calibration sets: for one fixed set, the miss rate is itself random, centred near $\alpha$, with a standard deviation of \cfConfPosthocCalsd{} points (Sec.~\ref{sec:cert}). The spread $\sigma_{B,l}$ is floored at $10^{-3}$
so that a confident prediction cannot make its margin vanish.

\section{Pre-registration and disclosures}\label{app:prereg}
This appendix lets a reader judge the evidence in Sec.~\ref{sec:exp}: what was fixed in advance, what changed, and
where the numbers need care.

\emph{Fixed in advance:} the claims and pass/fail bars; $\alpha=0.10$ and $U^\ast=1\,\%$; seven main comparisons
P1--P7 with Holm correction; an episode bootstrap (assembly pairs for merges) with 2{,}000 resamples; and a power rule
(a per-decision bar counts as powered only with at least 20{,}000 incapable decisions). \emph{Amendments before
scoring:} the episode bar uses all decisions; the model checkpoint was fixed; index ranges were trimmed to what compute
allowed; per-kind bars were added for merges, multi-module children and lone modules; and a cell that exceeds its bar
while underpowered counts as failed. The run was scored once.

\emph{Disclosures.} (i) The P1/P2 scorer used each method's own episodes; restricted post hoc to the \cfConfPosthocNcommon{} shared episodes, the gains are \cfConfPosthocPOne{} and \cfConfPosthocPTwo{} points. The matched point did not hold its 1\,\% target (\cfConfMainEdecMatched{} overall; merges \cfConfMainKindMergeMatched, multi-module children \cfConfMainKindMultiMatched, lone modules \cfConfMainKindLoneMatched), so realised risks differ between methods there; the own-bound comparison is the primary one. The stricter all-bounds-correct event of Eq.~\eqref{eq:guarantee} occurred in \cfConfPosthocMiscov{} of episodes (post hoc). (ii) Four planned comparison methods (raw
posterior, data-only bound, conformal selection, simulator-likelihood sequential Monte Carlo) have no rows in this run.
(iii) Our 100 calibration episodes came from the training pipeline, not the planned calibration split, and the
calibration range was trimmed after its labels existed; all confidence intervals are conditional on that set. (iv) In
the merge study only fault modes are inferred; other variations are known, and the merge margins were fitted on earlier
exploratory merges. (v) In the motor-model study the physics model knows the fault modes; the calibrated margin on T2, fitted at $s=0$, certifies few capable decisions even on dominated directions, which suggests a construction error; the data-only bound's margin was pooled over all severities.
(vi) Failed bars not discussed in the main text: our certificate against a direct capability predictor on low-grip
rollers (difference in certified-share drop, direct minus ours: \cfConfHetCfdirectHone{} points), and the precision
(\cfConfExplainDeficit) and false-reason rate (\cfConfExplainFalsereason) of natural-language explanations of
refusals; only the lone-module per-decision cell met the power rule. (vii) Fig.~\ref{fig:teaser} is illustrative: among
\cfTeaserPool{} qualifying test episodes, a fixed rule picks the one that certifies the most configurations with no
unsafe certification and all three verdicts. (viii) The extra-test results (App.~\ref{app:explore}) come from T0; each
bar was written before its run on fresh seeds, but the sequence of methods tried was adaptive.

\section{Extra-test study (exploratory)}\label{app:explore}
This appendix gives the setting and full table behind Sec.~\ref{sec:extra}.
The decision-layer results of Sec.~\ref{sec:extra} and Table~\ref{tab:probe} use T0 with knee motor models of severity
0--3, posteriors over fault modes and motor model, and an envelope bound calibrated end to end. Jobs come from a library
of realistic tasks; utility counts served jobs (120\,s each), unsafe commitments ($-600$\,s), and test and split
time. The hard screen kept the hypotheses whose residual on the test-drive segments is below a fixed threshold; the calibrated
screen sets that threshold by split-conformal on exploratory knee episodes (\cfScreenNominalNeps, \cfScreenKneeNeps{}
and \cfScreenHosNeps{} test episodes for the nominal, calibrated and held-out-severity arms, on different fresh seed
sets). Decision-EVOI ran the extra test in \cfProbeShareEvoi{} of episodes. Replacing the hard hypothesis screen by one
whose threshold is calibrated on misspecified-law episodes raised the share of jobs served from
\cfScreenNominalUseful{} to \cfScreenKneeUseful, with \cfScreenKneeEep{} unsafe episodes. Calibrated on severities
0--2 only and tested at severity 3, it served \cfScreenHosUseful{} with \cfScreenHosEep{} unsafe episodes.

\begin{table}[h]\centering\small
\caption{Testing when the motor model may be wrong (exploratory, T0, \cfScreenKneeNeps{} fresh episodes, knee laws of severity 0--3).
Useful: share of oracle-served jobs that the policy serves. Regret relative to the oracle and extra test time beyond
the 12-s standard test drive, in seconds.}\label{tab:probe}
\begin{tabular}{@{}lrrrr@{}}\toprule
Policy & Useful & $E_{\mathrm{unsafe}}$ & Regret & Probe\\\midrule
Standard drive only & \cfTFourUsefulFixedVTwo & \cfTFourEpunsafeFixedVTwo & \cfTFourRegretFixedVTwo & 0\\
Always clamp & \cfTFourUsefulFixedVTwoClamp & \cfTFourEpunsafeFixedVTwoClamp & \cfTFourRegretFixedVTwoClamp & 8.8\\
Dominance trigger & \cfTFourUsefulDominanceClamp & \cfTFourEpunsafeDominanceClamp & \cfTFourRegretDominanceClamp & \cfProbeExtraDominance\\
\textbf{Decision-EVOI} & \cfTFourUsefulEvoi & \cfTFourEpunsafeEvoi & \textbf{\cfTFourRegretEvoi} & \textbf{\cfProbeExtraEvoi}\\
\bottomrule\end{tabular}
\end{table}
When should the robot test before deciding? In T0 with uncertain motor models (exploratory, replicated on three
disjoint seed sets), choosing the clamp test by its expected value for the decision served as many jobs as always
testing (\cfTFourUsefulEvoi{} against \cfTFourUsefulFixedVTwoClamp) at a fraction of the test time
(Table~\ref{tab:probe}). The change in regret against always testing was \cfTFourDregretEvoiClamp\,s (95\,\% CI
\cfTFourDregretloEvoiClamp{} to \cfTFourDregrethiEvoiClamp), and \cfTFourDregretEvoiDominance\,s against probing
whenever a needed point is undominated. The gain is test time saved, not better decisions given the same information.
\end{document}